\pdfoutput=1

\documentclass[11pt]{article}

\usepackage[]{EMNLP2022}

\usepackage{times}
\usepackage{latexsym}

\usepackage[T1]{fontenc}

\usepackage[utf8]{inputenc}

\usepackage{microtype}

\usepackage{inconsolata}

\usepackage{amsmath}
\usepackage{amssymb}
\usepackage{amsfonts}
\usepackage{color}
\usepackage{makecell}
\usepackage{multirow}
\usepackage{graphicx}
\graphicspath{{figures/}}
\definecolor{my_blue}{RGB}{77, 115, 190}

\title{Q-TOD: A Query-driven Task-oriented Dialogue System}

\author{
    Xin Tian\thanks{~~Equal contribution.} \quad
    Yingzhan Lin\footnotemark[1] \quad
    Mengfei Song\footnotemark[1] \quad
    Siqi Bao \\
    \textbf{Fan Wang \quad
    Huang He \quad
    Shuqi Sun \quad
    Hua Wu} \\
    Baidu Inc., China \\
    \texttt{\{tianxin06, linyingzhan01, songmengfei01\}@baidu.com}
}

\begin{document}
\maketitle

\begin{abstract}
    Existing pipelined task-oriented dialogue systems usually have difficulties adapting to unseen domains, whereas end-to-end systems are plagued by large-scale knowledge bases in practice.
    In this paper, we introduce a novel query-driven task-oriented dialogue system, namely Q-TOD.
    The essential information from the dialogue context is extracted into a query, which is further employed to retrieve relevant knowledge records for response generation.
    Firstly, as the query is in the form of natural language and not confined to the schema of the knowledge base, the issue of \textit{domain adaption} is alleviated remarkably in Q-TOD.
    Secondly, as the query enables the decoupling of knowledge retrieval from the generation, Q-TOD gets rid of the issue of \textit{knowledge base scalability}.
    To evaluate the effectiveness of the proposed Q-TOD, we collect query annotations for three publicly available task-oriented dialogue datasets.
    Comprehensive experiments verify that Q-TOD outperforms strong baselines and establishes a new state-of-the-art performance on these datasets.
\end{abstract}

\section{Introduction}

Task-oriented dialogue systems are designed to help users achieve their goals, such as restaurant reservation, calendar scheduling, and movie recommendation.
Typically, these systems need to rely on external knowledge bases to retrieve necessary information for response generation~\citep{eric-etal-2017-key,wen-etal-2017-network,eric-etal-2020-multiwoz}.
Some end-to-end trainable approaches try to encode the knowledge base into a memory module and attend relevant knowledge records for response generation~\citep{wu2019global,qin-etal-2020-dynamic,raghu-etal-2021-constraint}.
Since these end-to-end approaches need to continually refresh the memory module, a large-scale knowledge base will lead to a heavy computation burden and difficult joint optimization.
Recently, some works leverage the power of pre-trained language models and take the entire linearized knowledge base as the input to assist response generation~\citep{gou-etal-2021-contextualize,tianbao2022unifiedskg}.
However, the input sequence could easily become too long to feed into the transformer network.
Considering there are thousands or millions of records in industrial knowledge bases, the \textit{knowledge base scalability} becomes a critical challenge for these end-to-end approaches.

In these circumstances, the practical deployed systems tend to employ pipelined designs and strip out the component of knowledge retrieval~\citep{wen-etal-2017-network,NEURIPS2020_e9462095,su-etal-2022-multi}.
These pipelined systems usually consist of natural language understanding, dialogue state tracking, dialogue policy learning, and system response generation.
In order to retrieve relevant information from the external knowledge base, these approaches need to pre-define the schema of dialogue states according to the knowledge base.
Due to this kind of strong association, these pipelined systems have difficulties adapting to unseen domains, i.e., weak ability on \textit{domain adaptation}.

\begin{figure*}[thp]
    \setlength{\belowcaptionskip}{-6pt}
	\centering
	\includegraphics[width=0.98\textwidth]{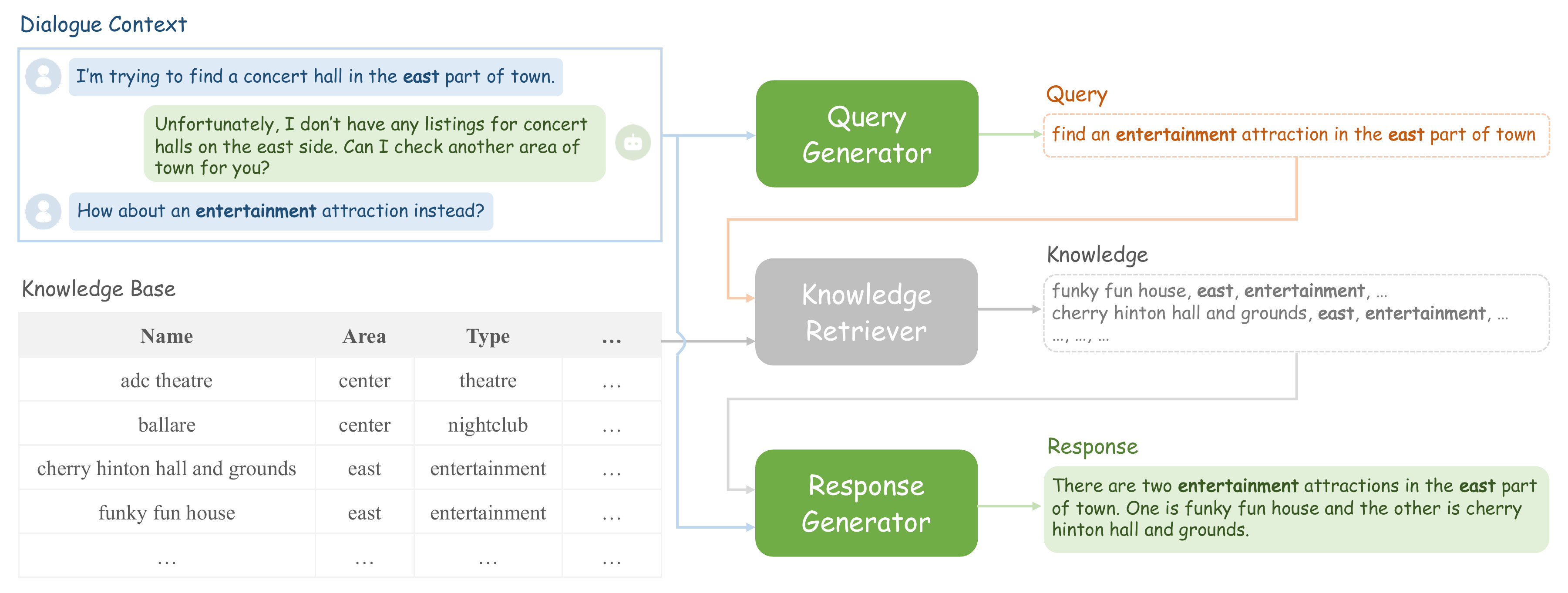}
	\caption{The overview of the proposed Q-TOD. Q-TOD consists of three modules, which are invoked sequentially: query generator, knowledge retriever, and response generator. Query generator and response generator are trained with a shared transformer, whereas knowledge retriever is an off-the-shelf retrieval model, allowing plug-and-play modularity.}
	\label{fig:architecture}
\end{figure*}

To tackle these issues, in this paper, we introduce a novel Query-driven Task-oriented Dialogue (Q-TOD) system.
The overview of Q-TOD is shown in Figure~\ref{fig:architecture}, where three sequential modules are included:
1) the \textit{query generator} extracts the essential information from the dialogue context into a concise query in an unstructured format of the natural language;
2) the generated query is then utilized to retrieve relevant knowledge records with an off-the-shelf \textit{knowledge retriever};
3) the \textit{response generator} produces a system response based on the retrieved knowledge records and the dialogue context.

The advantages brought by the query-driven task-oriented dialogue system are two-fold.
Firstly, the query is in the unstructured format of natural language, which is not confined to the knowledge base and is able to mitigate the issue of \textit{domain adaptation}.
Secondly, with the incorporation of the query, Q-TOD decouples the knowledge retrieval from the response generation, getting rid of the issue of the \textit{knowledge base scalability}.
To explore the effectiveness of the query-driven systems, we collect query annotations for three public task-oriented dialogue datasets: SMD~\citep{eric-etal-2017-key}, CamRest~\citep{wen-etal-2017-network}, and MultiWOZ-2.1~\citep{eric-etal-2020-multiwoz}.
Experimental results demonstrate that Q-TOD achieves superior performance as compared to other state-of-the-art approaches.
Particularly, in the few-shot settings, Q-TOD achieves a comparable performance with the previous state-of-the-art using only 5\% of the training data.
Our collected data, code, and models will be released at GitHub\footnotemark[1], hoping to facilitate further research in task-oriented dialogue systems.
\footnotetext[1]{\url{https://github.com/PaddlePaddle/Knover/tree/develop/projects/Q-TOD}}

\section{Methodology}

The goal of this paper is to explore a novel and effective framework for task-oriented dialogue systems.
As shown in Figure \ref{fig:architecture}, the proposed Q-TOD consists of three subsequent modules: query generator, knowledge retriever, and response generator.
The detailed design of these three modules will be discussed in the following.

\subsection{Query Generator}

The \textit{query generator} aims to extract essential information from the dialogue context into a natural language query.
For a multi-turn conversation, the dialogue context at the $t$-th turn can be represented as $C_t=\{U_0,R_0,\dots,U_t\}$, where each turn consists of user utterance $U_i$ and system response $R_i$.
With the dialogue context as input, the query at the $t$-th turn is generated with a Transformer-based language model: 
\begin{equation}
    Q_t=\textit{Transformer}\left(C_t\right)
\end{equation}
In the query generation, the noisy or out-of-date information from the context is supposed to be removed, whereas the essential and up-to-date requirements raised by the user should be highlighted.
As shown in the example of Figure \ref{fig:architecture}, the query only contains the minimal user requirements in the current turn (i.e., find an east entertainment attraction) and discards the outdated attraction type of concert hall.
In cases where no query is required, e.g. greetings or thanks, a special token \texttt{[NOTHING]} is used to represent the null query at this turn.

Recently, in some knowledge-intensive conversations, there is a trend to employ the query to enhance the performance of relevant knowledge retrieval.
In conversational question answering, to deal with ellipsis and coreference, a question rewriting task is introduced to convert a context-dependent question into a self-contained query~\citep{vakulenko2021question,anantha-etal-2021-open}.
In open-domain knowledge-grounded dialogue, to incorporate real-time external information, some works learn to generate a search query and leverage search engines for response generation~\citep{Komeili2022InternetAugmentedDG,Shuster2022LanguageMT}.
To the best of our knowledge, Q-TOD is the first work that tries to encode the natural language query into a task-oriented dialogue system.
Distinct from the above approaches, the query in Q-TOD is designed to extract the essential and up-to-date user requirements.

\subsection{Knowledge Retriever}

The \textit{knowledge retriever} utilizes the generated query to retrieve relevant knowledge records from the external knowledge base:
\begin{equation}
    K_t=\textit{Retriever}\left(Q_t;\mathcal{K}\right)
\end{equation}
where $\mathcal{K}$ refers to the entire knowledge base, and $K_t=\{k_t^1,k_t^2,\dots,k_t^n\}$ are retrieved top-$n$ relevant knowledge records.
As displayed in Figure~\ref{fig:architecture}, the module of knowledge retriever is a black box in this system.
In fact, any off-the-shelf knowledge retriever can be employed in Q-TOD, including BM-25 or dense retrieval models.
Such strong adaptability and flexibility mainly benefit from the preceding query generation.
Firstly, the query is in the format of natural language, which is a universal representation and adaptable to commonly used retrievers.
Secondly, although multi-turn dialogue context typically requires elaborately designed or tuned retrievers \citep{shuster-etal-2021-retrieval-augmentation}, by extracting essential information into a concise query, the off-the-shelf retriever can achieve relatively good performance as well.

In fact, the module of knowledge retriever is the key to the knowledge base scalability of Q-TOD.
Given a large-scale knowledge base, the retriever filters out massive irrelevant knowledge records and picks out top-$n$ relevant ones.
In this way, the subsequent response generation is not affected by the size of the knowledge base and is able to pay more attention to knowledge utilization.
As suggested by recent works~\citep{Adolphs2021ReasonFT,Shuster2022LanguageMT} and verified in our experiments, the decoupling of knowledge retrieval and response generation alleviates the modeling difficulty and boosts the final performance.

\subsection{Response Generator}

The \textit{response generator} produces the system response given the retrieved top-$n$ knowledge records and dialogue context:
\begin{equation}
    R_t=\textit{Transformer}\left(K_t;C_t\right)
\end{equation}
With the assistance of the preceding modules, the response generator can focus more on precise knowledge utilization and produce high-quality replies towards the dialogue context. 
~\\

In Q-TOD, we train the query generator and response generator jointly with a shared transformer.
To distinguish these tasks in a single model, two task-specific discrete prompts $Z_Q$ and $Z_R$ are adopted and concatenated with the rest input.
For query generation, the prompt $Z_Q$ is ``\texttt{translate dialogue context to query:}''.
For response generation, the prompt $Z_R$ is ``\texttt{generate system response based on knowledge and dialogue context:}''.
Overall, the training objective is to minimize the following negative log-likelihood loss:
\begin{align}
    \nonumber \mathcal{L}=&-\log P_\Theta\left(Q_t|Z_Q;C_t\right)\\
    &-\log P_\Theta\left(R_t|Z_R;K_t;C_t\right)
\end{align}
The knowledge retrieval is a black box in this system and thus not involved in the optimization.

\section{Experiments}

\subsection{Datasets}

\begin{table}[t]
    \renewcommand\arraystretch{1.2}
    \centering
    \small
    \begin{tabular}{lccc}
        \hline
        \textbf{Statistics} & \textbf{SMD} & \textbf{CamRest} & \textbf{MWOZ} \\
        \hline
        \hline
        Dialogues & 3031 & 676 & 2097 \\
        Utterances & 15928 & 5488 & 19632 \\
        Domains & 3 & 1 & 3 \\
        Turns per Dialogue & 5.26 & 8.12 & 8.89 \\
        Tokens per Utterance & 7.97 & 12.31 & 14.73 \\
        Tokens per Query & 7.50 & 9.67 & 10.57 \\
        \hline
    \end{tabular}
    \caption{Statistics of the datasets in the experiments.}
    \label{tab:dataset}
\end{table}

\begin{table*}[th]
    \setlength{\belowcaptionskip}{-6pt}
    \centering
    \begin{tabular}{lcccccc}
        \hline
        \multirow{2}*{\textbf{Model}} & \multicolumn{2}{c}{\textbf{SMD}} & \multicolumn{2}{c}{\textbf{CamRest}} & \multicolumn{2}{c}{\textbf{MWOZ}} \\
         & \textbf{Entity F1} & \textbf{BLEU} & \textbf{Entity F1} & \textbf{BLEU} & \textbf{Entity F1} & \textbf{BLEU} \\
        \hline
        \hline
        DSR & 51.90$^\dag$ & 12.70$^\dag$ & 53.60$^\dag$ & 18.30$^\dag$ & 30.00$^\ddag$ & 9.10$^\ddag$ \\
        KB-Retriever & 53.70 & 13.90 & 58.60 & 18.50 & - & - \\
        GLMP & 60.70$^\ddag$ & 13.90$^\ddag$ & 58.90$^\S$ & 15.10$^\S$ & 32.40$^\ddag$ & 6.90$^\ddag$ \\
        DF-Net & 62.70 & 14.40 & - & - & 35.10 & 9.40 \\
        GPT-2+KE & 59.78 & 17.35 & 54.85 & 18.00 & 39.58 & 15.05 \\
        CDNET & 62.90 & 17.80 & 68.60 & 21.80 & 38.70 & 11.90 \\
        COMET & 63.60 & 17.30 & - & - & - & - \\
        UnifiedSKG (T5-Large) & 65.85 & 17.27 & 71.03$^\ast$ & 20.31$^\ast$ & 46.04$^\ast$ & 13.69$^\ast$ \\
        UnifiedSKG (T5-3B) & 67.88 & 15.45 & 72.78$^\ast$ & 18.46$^\ast$ & 49.65$^\ast$ & 13.01$^\ast$ \\
        \hline
        Q-TOD (T5-Large) & 71.11 & 21.33 & 74.22 & 23.75 & 50.61 & 17.62 \\
        Q-TOD (T5-3B) & \textbf{73.44} & \textbf{21.76} & \textbf{76.81} & \textbf{24.65} & \textbf{53.28} & \textbf{18.27} \\
        \hline
    \end{tabular}
    \caption{Experimental results on the SMD, CamRest, and MWOZ datasets, with the best value written in bold~\footnotemark[3]. $\dag$, $\ddag$, $\S$ denotes that the results are cited from~\citet{qin-etal-2019-entity}, \citet{qin-etal-2020-dynamic}, and~\citet{raghu-etal-2021-constraint}, respectively. $\ast$ indicates that we reproduce the results using the official code released by the authors.}
    \label{tab:sota}
\end{table*}

To investigate the effectiveness of the proposed query-driven dialogue system, we collect query annotations for three publicly available multi-turn task-oriented dialogue datasets: Stanford Multi-Domain (SMD)~\citep{eric-etal-2017-key}, CamRest~\citep{wen-etal-2017-network}, and MultiWOZ-2.1 (MWOZ)~\footnotemark[2]~\citep{eric-etal-2020-multiwoz}.
The query annotations are collected through three stages.
The authors first provide ten examples of dialogue sessions with query annotations for each dataset.
Then, the crowd workers complete all query annotations on these dialogues after reading the examples.
Finally, to ensure the quality of annotations, multiple data specialists will review it.
We will release the collected data for further research.

\footnotetext[2]{For MultiWOZ-2.1, following previous works~\citep{qin-etal-2020-dynamic, raghu-etal-2021-constraint}, we use the version released by~\citet{qin-etal-2020-dynamic}, which equips each dialogue with corresponding knowledge base.}

In our experiments, we utilize the provided training/validation/test partitions of all benchmark datasets.
Table~\ref{tab:dataset} summarizes the statistics of the above three datasets.

\subsection{Experimental Settings}

Our experiments are carried out with T5~\citep{JMLR:v21:20-074}, in which two model sizes are used: T5-Large and T5-3B.
For knowledge retriever, we leverage an off-the-shelf retrieval model RocketQA~\citep{ren-etal-2021-rocketqav2}.
Particularly, we fine-tune T5 with AdamW optimizer~\citep{DBLP:conf/iclr/LoshchilovH19} and Noam learning rate scheduler~\citep{vaswani2017attention}.
During inference, the decoding strategy of beam search is employed, with a beam size of 4.
And the number of retrieved knowledge records top-$n$ is set to 3.
All the models are trained on 8 NVIDIA Tesla A100 GPU cards for 50 epochs and early stopped according to the performance on the validation set.
More details about hyper parameter settings are provided in Appendix~\ref{sec:appendix-hyper-parameters}.

\subsection{Baselines}

We compare Q-TOD with the following strong baselines:

\noindent\textbf{DSR}~\citep{wen-etal-2018-sequence} models dialogue state as distributed representation to query the knowledge base with an attention mechanism.

\noindent\textbf{KB-Retriever}~\citep{qin-etal-2019-entity} proposes an entity-consistency augmented decoder to focus on a single row of the knowledge base by memory network and attention mechanism.

\noindent\textbf{GLMP}~\citep{wu2019global} leverages a global-to-local pointer network to first generate a sketch response and then fill slots with entities from the knowledge base.

\noindent\textbf{DF-Net}~\citep{qin-etal-2020-dynamic} applies the Mixture-of-Experts mechanism (MoE) to dynamically exploit the relevance between the target domain and all source domains.

\noindent\textbf{GPT-2+KE}~\citep{madotto-etal-2020-learning} proposes to pack the knowledge base into the model parameters implicitly through dialogue data augmentation.

\noindent\textbf{CDNET}~\citep{raghu-etal-2021-constraint} computes a distillation distribution over the knowledge records, which is used to get the final copy distribution for entity choosing.

\noindent\textbf{COMET}~\citep{gou-etal-2021-contextualize} introduces a Memory-Masked Encoder to enforce entities interact within the same knowledge record, aiming to avoid the distraction from the irrelevant ones.

\noindent\textbf{UnifiedSKG}~\citep{tianbao2022unifiedskg} recasts 21 structured knowledge grounding tasks into a unified text-to-text language model (including task-oriented dialogue modeling) and achieves state-of-the-art performance on these tasks.

\subsection{Results}

\footnotetext[3]{Since UnifiedSKG includes a base size version on SMD, we also train a T5-Base Q-TOD for comparison. Q-TOD (T5-Base) obtains 68.22\% Entity-F1 and 20.14\% BLEU, which outperforms UnifiedSKG on the same scale (UnifiedSKG T5-Base, 66.45\% Entity-F1, 17.41\% BLEU).}

Following the prior works~\citep{eric-etal-2017-key,wu2019global,qin-etal-2020-dynamic,madotto-etal-2020-learning,raghu-etal-2021-constraint,gou-etal-2021-contextualize,tianbao2022unifiedskg}, we evaluate the model performance with metrics of micro \textbf{Entity-F1}~\citep{eric-etal-2017-key} and \textbf{BLEU}~\citep{papineni-etal-2002-bleu}.
The Entity-F1 measures the model's abilities to generate relevant entities from the external knowledge base according to the dialogue context.
BLEU measures the n-gram overlap between generated response and the oracle response.

Table~\ref{tab:sota} summarizes the results of Q-TOD and all baselines on three datasets.
It can be observed that our Q-TOD consistently outperforms all previous models, achieving a new state-of-the-art result.
Specifically, on the Entity-F1 metric, Q-TOD achieves the absolute improvement of 5.56\% on SMD, 4.03\% on CamRest, and 3.36\% on MWOZ, respectively.
On the BLEU metric, Q-TOD also obtains the highest score with the increment of 3.96\%, 2.85\%, and 3.22\% on SMD, CamRest, and MWOZ, respectively~\footnotemark[4].
These results demonstrate that the proposed Q-TOD can generate high-quality system responses with the relevant knowledge records.
It is worth noting that, UnifiedSKG~\citep{tianbao2022unifiedskg} employs a Transformer-based response generator similar to ours, but our model surpasses it on all three datasets under both T5-Large and T5-3B sizes.
This confirms the benefits of our proposed query-driven retrieval, which helps the response generator avoid distractions from irrelevant knowledge records and focus on the utilization of relevant ones.

\footnotetext[4]{The three datasets we used include only one reference system response, which results in comparably low BLEU scores for Q-TOD and all baselines.}

\begin{figure}[t]
    \setlength{\abovecaptionskip}{3pt}
    \setlength{\belowcaptionskip}{-6pt}
	\centering
	\includegraphics[width=0.45\textwidth]{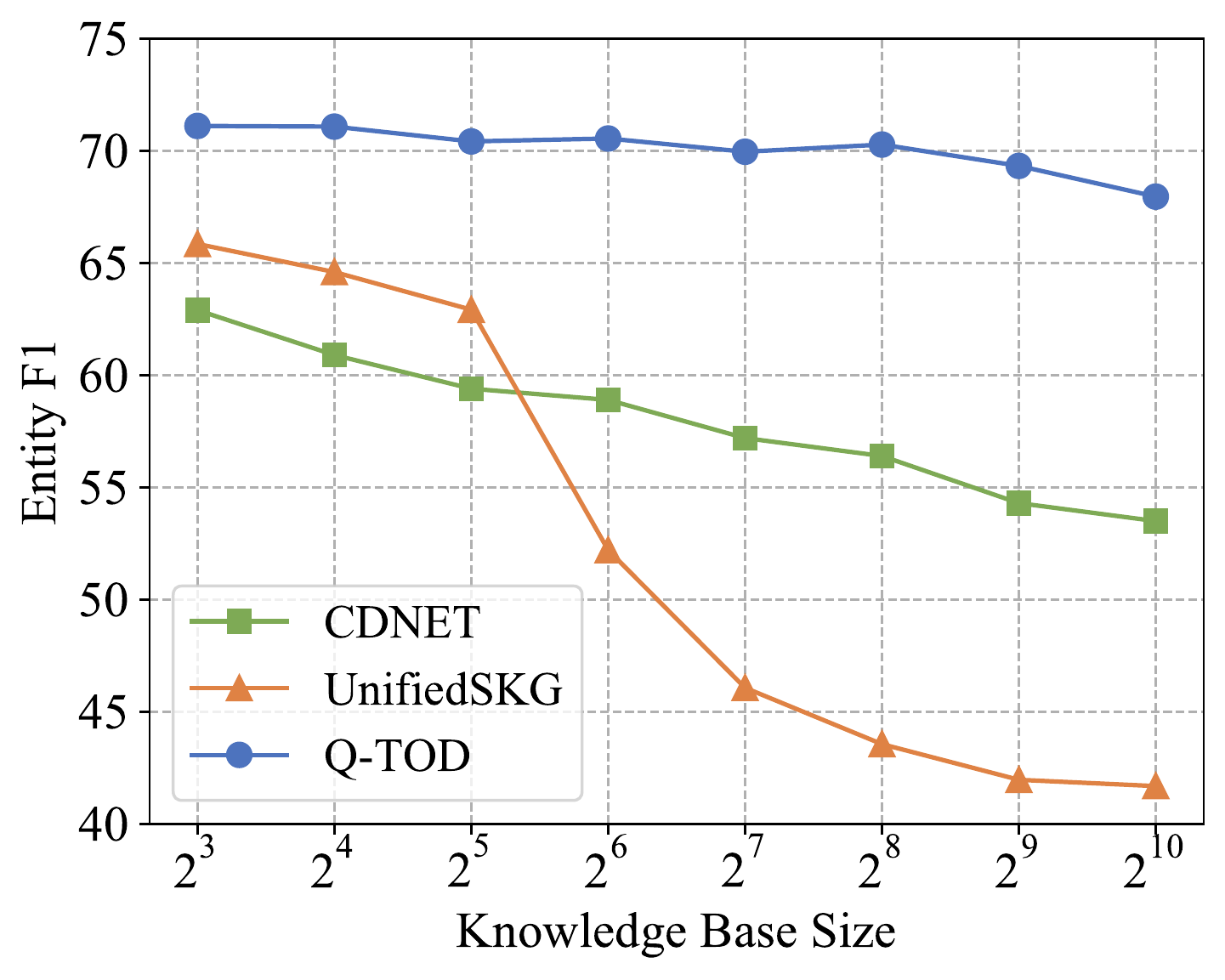}
	\caption{Effects of knowledge base scaling.}
	\label{fig:knowledge-base-scalability}
\end{figure}

\section{Discussion}

For further analysis of Q-TOD, we will have discussions on the following aspects: knowledge base scalability, effect of query, performance of precise knowledge, domain adaptation, and case study.
In this section, unless specified, experiments are carried out with T5-Large.

\subsection{Knowledge Base Scalability}

In practice, the knowledge base of a specific domain usually contains thousands of records, such as weather forecasts and music recommendations.
Hence, it is necessary to explore the knowledge base scalability in task-oriented dialogue systems.
To this end, we simulate large knowledge bases by expanding the existing ones.
Specifically, we expand the original session-level knowledge bases on the SMD dataset by injecting dataset-level knowledge records and some crawled knowledge records~\footnotemark[5].
\footnotetext[5]{The average size of the session-level knowledge base for SMD is 7.19. We construct a dataset-level knowledge base by merging all session-level knowledge bases, where the average size of the dataset-level knowledge base is 542.67. In our experiments, extra crawled knowledge records are added to the knowledge base when necessary.}

As shown in Figure~\ref{fig:knowledge-base-scalability}, we compare Q-TOD with two strong baselines, CDNET~\citep{raghu-etal-2021-constraint} and UnifiedSKG~\citep{tianbao2022unifiedskg}.
The results show that when increasing the knowledge base size from $2^3$ to $2^{10}$, Q-TOD is able to maintain a stable performance on Entity F1.
In particular, when the size of the knowledge base is expanded to $2^{10}$ (128 times compared with the original SMD), Q-TOD obtains 67.96\% Entity F1, only a decrease of 3.15\%.
The superior performance is owing to the fact that Q-TOD extracts the essential information from the dialogue context into the query.
The short query enables the knowledge retrieval to be decoupled from the response generation, getting rid of the issue of the \textit{knowledge base scalability}.
In contrast, the performance of CDNET and UnifiedSKG decreases gradually as the size of the knowledge base increases.
This indicates that joint modeling can barely adapt to large-scale knowledge bases, which might result from the difficulty of a single model in handling implicit knowledge retrieval and response generation simultaneously.
Moreover, due to the limitation of the max input length in UnifiedSKG, the Entity F1 decreases sharply when the size of the knowledge base is greater than $2^6$.

\subsection{Effect of Query}
\label{sec:effect-of-query}

To investigate the effectiveness of query incorporation, we would like to answer the following two research questions (RQ).

\begin{table}[tp]
    \setlength{\belowcaptionskip}{-6pt}
    \renewcommand\arraystretch{1.1}
    \centering
    \small
    \begin{tabular}{lcc}
        \hline
        \textbf{~~Model} & \textbf{~~~~~~~~~~Entity F1~~} \\
        \hline
        \hline
        ~~UnifiedSKG & ~~~~~~~~~~65.85~~ \\
        ~~Q$^I$-TOD & ~~~~~~~~~~69.57~~ \\
        ~~Q-TOD & ~~~~~~~~~~71.11~~ \\
        \hline
    \end{tabular}
    \caption{Comparison of performance between UnifiedSKG, Q$^I$-TOD, and Q-TOD on SMD.}
    \label{tab:qitod}
\end{table}

\begin{figure}[t]
    \setlength{\abovecaptionskip}{3pt}
    \setlength{\belowcaptionskip}{-6pt}
	\centering
	\includegraphics[width=0.45\textwidth]{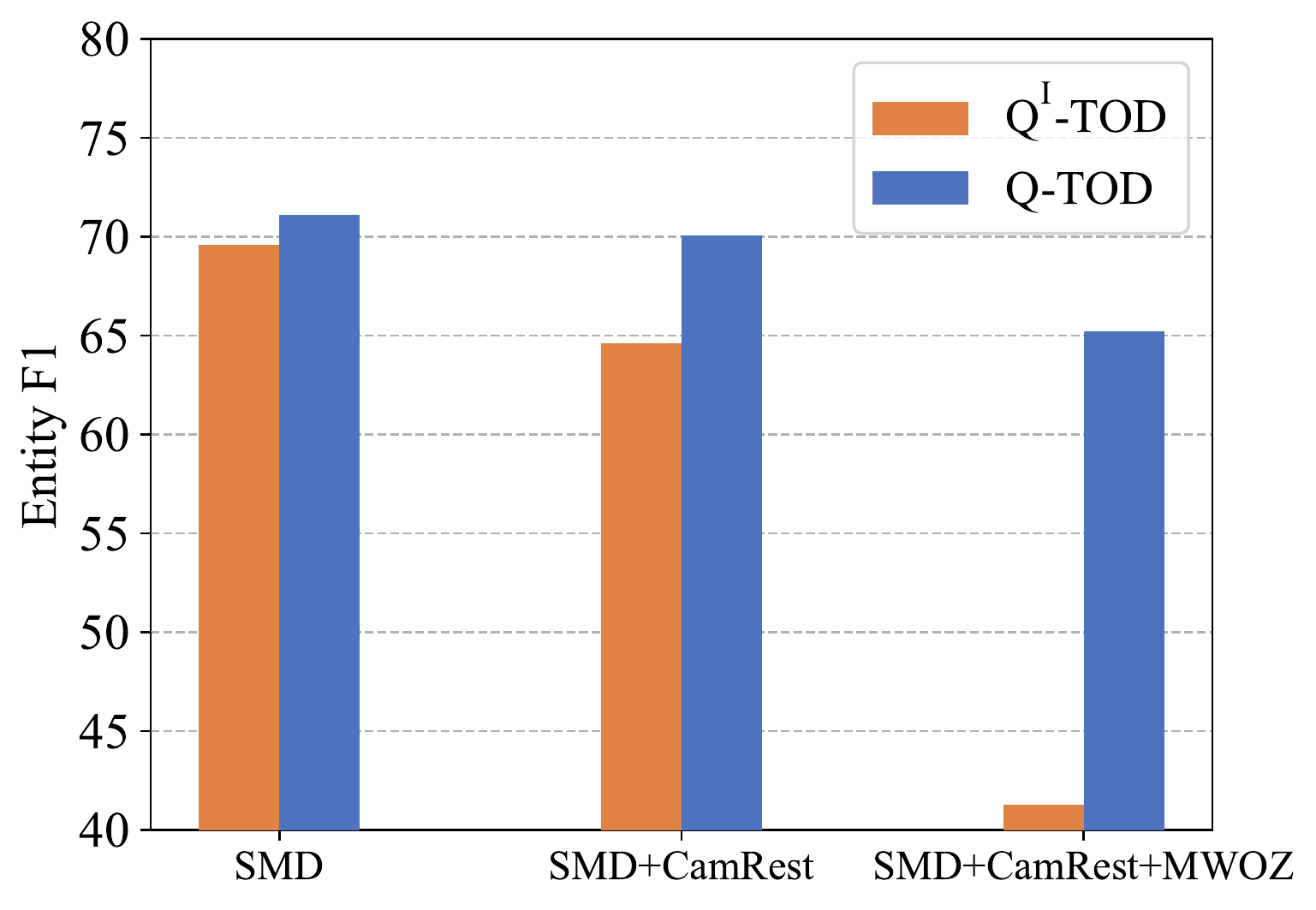}
	\caption{Comparison of performance between Q$^I$-TOD and Q-TOD.}
	\label{fig:effect-of-query}
\end{figure}

\noindent \textbf{RQ1: }\textit{what would happen to Q-TOD without query annotations?}

Under our framework, to deal with the situation without any query annotation, a straightforward solution is to degrade the query generator to identical mapping.
In other words, the dialogue context itself can be regarded as a naive query to retrieve knowledge records for response generation.
This setting with identical mapping is denoted as Q$^I$-TOD. Experiments with Q$^I$-TOD are carried out on SMD and results are summarized in Table~\ref{tab:qitod}.
Q$^I$-TOD obtains 69.57\% Entity F1, outperforming the previous state-of-the-art UnifiedSKG (65.85\% Entity F1).
This indicates that our framework can also achieve state-of-the-art even without query annotations.

\noindent \textbf{RQ2: }\textit{why not content with Q$^I$-TOD?}

In practical deployments, the conversations are more complex and noisy as compared to those in public datasets.
For instance, one user might ask for navigation after a restaurant reservation, also known as a cross-domain conversation.
Since the context contains distractions of the noisy or outdated information, Q$^I$-TOD encounters difficulties in retrieving relevant knowledge records.
For better comparison between Q$^I$-TOD and Q-TOD, we further construct two cross-domain dialogue datasets by merging dialogue sessions from SMD, CamRest, and MWOZ.
Detailed construction process is described in Appendix~\ref{sec:appendix-cross-domain-dataset-construction}.
As shown in Figure~\ref{fig:effect-of-query}, the gap between Q$^I$-TOD and Q-TOD becomes larger on cross-domain datasets, especially more than 20\% Entity-F1 on \texttt{SMD+CamRest+MWOZ}.
The limited performance of knowledge retrieval of Q$^I$-TOD results in low-quality system responses in cross-domain scenarios.
These results suggest that the query generator is a crucial module in our framework, which ensures that the noisy or out-of-date information is filtered out and will not be transmitted to the knowledge retriever.

\subsection{Performance of Precise Knowledge}

\begin{table}[tp]
    \renewcommand\arraystretch{1.1}
    \centering
    \small
    \begin{tabular}{lc}
        \hline
        \textbf{~~Model} & \textbf{~~~~~Entity F1~~} \\
        \hline
        \hline
        ~~Q-TOD (T5-Large) & ~~~~~71.11~~ \\
        \quad ~~w/ fine-tuned retriever & ~~~~~71.17 (+0.06)~~ \\
        \quad ~~w/ oracle knowledge & ~~~~~71.96 (\textbf{+0.85})~~ \\
         \hline
        ~~Q-TOD (T5-3B) & ~~~~~73.44~~ \\
        \quad ~~w/ fine-tuned retriever & ~~~~~74.96 (+1.52)~~ \\
        \quad ~~w/ oracle knowledge & ~~~~~76.20 (\textbf{+2.76})~~ \\
        \hline
    \end{tabular}
    \caption{Performance of more precise knowledge. \textit{Oracle knowledge} refers to taking the ground-truth of knowledge as the input of the response generator, which is the theoretical upper bound.}
    \label{tab:precise-knowledge}
\end{table}

Although the off-the-shelf knowledge retriever is utilized in Q-TOD, a domain-specific knowledge retriever can be employed to obtain precise knowledge when the knowledge annotation is available.
To investigate the effect of precise knowledge, we collect turn-level knowledge annotations to fine-tune a new knowledge retriever on the SMD dataset.
Here, the knowledge retriever is initialized with T5~\citep{JMLR:v21:20-074}, where the model takes the query and each linearized knowledge record as input and outputs a relevance label: \texttt{MATCHED} or \texttt{MISMATCHED}.

The results are summarized in Table~\ref{tab:precise-knowledge}.
It can be observed that the system with fine-tuned knowledge retriever achieves better Entity-F1 than the off-the-shelf retriever.
To give an idea of the performance limits of knowledge retrieval, we also evaluate it with oracle knowledge records.
The performance with oracle knowledge shows that there is some headroom for Q-TOD when improving the knowledge precision.
Moreover, we note that compared with T5-Large, T5-3B achieves more improvements using more precise knowledge, especially the improvement of Entity F1 becomes 2.76\% with oracle knowledge.
From the insights of previous works~\citep{thoppilan2022lamda}, this might thank to a greater ability of knowledge utilization in a larger model.

\begin{table*}[th]
    \renewcommand\arraystretch{1.3}
    \centering
    \small
    \begin{tabular}{lcccccccc}
        \hline
        \multirow{2}*{\textbf{Model}} & \multicolumn{2}{c}{\textbf{Zero-Shot}} & \multicolumn{2}{c}{\textbf{Few-Shot 1\%}} & \multicolumn{2}{c}{\textbf{Few-Shot 5\%}} & \multicolumn{2}{c}{\textbf{Few-Shot 20\%}} \\
         & \textbf{Entity F1} & \textbf{BLEU} & \textbf{Entity F1} & \textbf{BLEU} & \textbf{Entity F1} & \textbf{BLEU} & \textbf{Entity F1} & \textbf{BLEU} \\
        \hline
        \hline
        PPTOD$^\dag$ & 10.24 & 0.73 & 19.98 & 6.19 & 23.93 & 7.92 & 29.64 & 9.79 \\
        UnifiedSKG$^\dag$ & 28.62 & 7.61 & 36.66 & 10.70 & 41.57 & 12.72 & 45.37 & 13.04 \\
        \hline
        Q-TOD & \textbf{31.52} & \textbf{10.14} & \textbf{39.19} & \textbf{11.35} & \textbf{45.99} & \textbf{14.27} & \textbf{48.54} & \textbf{16.51} \\
        \hline
    \end{tabular}
    \selectfont
    \caption{Performance of zero-shot and few-shot settings. The average scores over three runs are reported. $^\dag$ denotes that we reproduce the results using the official code released by the authors.}
    \label{tab:domain-adaption}
\end{table*}

\begin{table*}
    \renewcommand\arraystretch{1.3}
    \centering
    \footnotesize
    \begin{tabular}{ll}
        \hline
        \multirow{4}*{\textbf{Dialogue Context}} & U: I'm looking for an \textcolor{my_blue}{\textbf{expensive}} eastern european place in the \textcolor{my_blue}{\textbf{south}}. \\
        & S: Unfortunately, I don't happen to have any listing that meets what you were looking for. \\
        & ~~~~Would you like me to try either a different area or type of restaurant? \\
        & U: How about a \textcolor{my_blue}{\textbf{chinese}} restaurant? \\
        \hline
        \textbf{Generated Query} & find an \textcolor{my_blue}{\textbf{expensive}} \textcolor{my_blue}{\textbf{chinese}} restaurant in the \textcolor{my_blue}{\textbf{south}} part of the city \\
        \hline
        \multirow{3}*{\textbf{Retrieved Knowledge}} & peking restaurant, \textcolor{my_blue}{\textbf{chinese}}, \textcolor{my_blue}{\textbf{south}}, \textcolor{my_blue}{\textbf{expensive}}, ... \\
        & the good luck chinese food takeaway, \textcolor{my_blue}{\textbf{chinese}}, \textcolor{my_blue}{\textbf{south}}, \textcolor{my_blue}{\textbf{expensive}}, ... \\
        & ugly duckling, \textcolor{my_blue}{\textbf{chinese}}, centre, \textcolor{my_blue}{\textbf{expensive}}, ... \\
        \hline
        \multirow{2}*{\textbf{Generated Response}} & There are two expensive chinese restaurants in the south part of town. Peking restaurant \\
        & and the good luck chinese food takeaway. \\
        \hline
        \multirow{2}*{\textbf{Oracle Response}} & Certainly, we have two expensive chinese restaurants the good luck chinese food takeaway,\\
        & and peking restaurant. Would you like a reservation at one, if so for how many? \\
        \hline
    \end{tabular}
    \caption{Cherry-picked case of Q-TOD. \textcolor{my_blue}{Blue} indicates the essential and updated information in the dialogue context.}
    \label{tab:cherry-picked-case}
\end{table*}

\begin{table*}[tp]
    \setlength{\abovecaptionskip}{8pt}
    \setlength{\belowcaptionskip}{-6pt}
    \renewcommand\arraystretch{1.3}
    \centering
    \small
    \begin{tabular}{ll}
        \hline
        \multirow{2}*{\textbf{Dialogue Context}} & U: I am looking for a place to stay. The hotel should have a star of 2 and should be in the \\
        ~ & moderate price range. \\
        \hline
        \textbf{Generated Query} & find a moderately priced 2 star hotel \\
        \hline
        \multirow{3}*{\textbf{Retrieved Knowledge}} & ashley hotel, north, moderate, 2 star, ... \\
        & lovell lodge, north, moderate, 2 star, ... \\
        & a and b guest house, east, moderate, 4 star, ... \\
        \hline
        \textbf{Generated Response} & There are 2 hotels that meet your criteria. \textcolor{red}{\textbf{Would you like to stay in the north or the centre?}} \\
        \hline
        \multirow{2}*{\textbf{Oracle Response}} & There are two such hotels in the north area. The first is the ashley hotel and the second is the \\
        ~ & lovell lodge. Do you have a preference? \\
        \hline
    \end{tabular}
    \caption{Lemon case of Q-TOD. \textcolor{red}{Red} refers to the factually incorrect statements in the generated system response.}
    \label{tab:lemon-case}
\end{table*}

\subsection{Domain Adaption}

A primary advantage of our framework is its strong ability on domain adaptation.
To this end, we compare Q-TOD with two strong baselines in zero/few-shot settings, PPTOD~\citep{su-etal-2022-multi} and UnifiedSKG~\citep{tianbao2022unifiedskg}.
PPTOD achieves the state-of-the-art performance in pipelined dialogue systems, which relies on the predefined schema to train a plug-and-play model.
UnifiedSKG is the previous state-of-the-art model in end-to-end dialogue systems, which takes the entire knowledge base as input for response generation.

\subsubsection{Zero-Shot Setting}

To investigate the performance of Q-TOD in zero-shot settings, we train the model on SMD and CamRest, and then evaluate the performance on the MWOZ test set.
As seen in Table~\ref{tab:domain-adaption}, Q-TOD significantly surpasses the two baselines on both Entity F1 and BLEU metrics.
For PPTOD, since the method depends on the predefined schema, the performance is poor on unseen domains.
Without any training data, Q-TOD exceeds the baseline DSR~\citep{wen-etal-2018-sequence} in the full-training setting (31.52\% vs. 30.00\% on Entity F1).
These results verify the strong ability on domain adaptation of the proposed framework.

\subsubsection{Few-Shot Setting}

To further explore the performance of Q-TOD with a small number of training samples, we evaluate it and baselines in few-shot settings.
Specifically, the following three steps are carried out: training on SMD and CamRest, training on partial MWOZ training set, and evaluating on MWOZ test set.
As shown in Table~\ref{tab:domain-adaption}, our framework consistently outperforms all baseline models.
This demonstrates that Q-TOD is capable of transferring knowledge from other domains and achieves better performance in low-resource scenarios.
Notably, with only 5\% of the training data, Q-TOD achieves a comparable performance with the previous state-of-the-art model UnifiedSKG~\citep{tianbao2022unifiedskg} (45.99\% vs. 46.06\% on Entity F1).

\subsection{Case Study}
\label{sec:case-study}

For case study, we select and present two examples of Q-TOD, including the retrieved knowledge records and generated system responses.

A cherry-picked case generated by Q-TOD is shown in Table~\ref{tab:cherry-picked-case}.
It can be observed that the generated query successfully extracts the essential and up-to-date user requirements.
With this query, the off-the-shelf retriever outputs relevant and precise knowledge records.
Then the response generator produces a high-quality system response based on the retrieved knowledge records and the dialogue context.

Table~\ref{tab:lemon-case} describes a lemon case of Q-TOD.
Despite that the retrieved knowledge records are correct, the response generator produces a factually incorrect system response.
The system provides an option to the user, which doesn't exist in fact.
This suggests that Q-TOD also suffers from the well-known problem of knowledge hallucination, where it generates plausible looking responses that are factually incorrect.

\section{Related Work}

In task-oriented dialogue systems, there is a trend to develop end-to-end trainable approaches to incorporate the external knowledge base for response generation~\citep{wen-etal-2018-sequence,qin-etal-2019-entity,wu2019global,qin-etal-2020-dynamic,madotto-etal-2020-learning,raghu-etal-2021-constraint,gou-etal-2021-contextualize,tianbao2022unifiedskg}.
Some works encode the entire knowledge base into a memory module and learn to attend to the relevant knowledge entities for response generation~\citep{wen-etal-2018-sequence,qin-etal-2019-entity,wu2019global,qin-etal-2020-dynamic,raghu-etal-2021-constraint}.
Recently, with the advances in the pre-trained language models, some works take the entire linearized knowledge base as the transformer input and generate the final system response directly~\citep{gou-etal-2021-contextualize,tianbao2022unifiedskg}.
Additionally, \citet{madotto-etal-2020-learning} proposes to store the knowledge base in the language model parameters implicitly through dialogue augmentation.
However, considering the knowledge base usually contains thousands of records in practice, these end-to-end trainable approaches face the critical challenges of knowledge base scalability.

Meanwhile, there are many pipelined task-oriented dialogue systems, which strip out the component of knowledge retrieval~\citep{young2013pomdp,wen-etal-2017-network,NEURIPS2020_e9462095,lin-etal-2020-mintl,Yang_Li_Quan_2021,Sun2022BORTBA,su-etal-2022-multi,he2022galaxy}.
They usually decompose a task-oriented dialogue system into several pipelined modules: natural language understanding, dialogue state tracking, dialogue policy learning, and system response generation~\citep{young2013pomdp,wen-etal-2017-network}.
Some recent works formulate all pipelined modules as a cascaded generation task using pre-trained language model~\citep{NEURIPS2020_e9462095,lin-etal-2020-mintl,Yang_Li_Quan_2021,peng-etal-2021-soloist,lee-2021-improving-end,Sun2022BORTBA}.
To further boost the performance, some models attempt to introduce the pre-training strategy into task-oriented dialogue systems~\citep{liu-etal-2021-pretraining,su-etal-2022-multi,he2022galaxy}.
However, these pipelined systems rely on the predefined schema to retrieve knowledge from an external knowledge base, leading to difficulties in adapting to unseen domains.

In other research fields, there are also some works using a query to store essential information or retrieve relevant knowledge.
In knowledge-grounded open-domain dialogue, to incorporate real-time external information, recent works learn to generate a search query based on the dialogue context for internet searching~\citep{Komeili2022InternetAugmentedDG,Adolphs2021ReasonFT,Shuster2022LanguageMT}.
In open-domain conversational question answering, to handle the reference problem and optimize retrieval performance, recent systems introduce a question rewriting task to convert a context-dependent question into a self-contained question~\citep{vakulenko2021question,anantha-etal-2021-open,wu2021conqrr}.
In context-dependent text-to-SQL task, some works learn to reformulate multi-turn conversational questions into a  self-contained question, and then a context-independent text-to-SQL parser follows~\citep{chen-etal-2021-decoupled,xiao2022cqr}.
To the best of our knowledge, Q-TOD is the first framework that introduces the query into task-oriented dialogue systems.

\section{Conclusion}

In this paper, we propose a novel query-driven task-oriented dialogue system, namely Q-TOD.
Q-TOD consists of three modules: the query generator extracts the essential and up-to-date information from the dialogue context into a concise query, the off-the-shelf knowledge retriever utilizes the generated query to retrieve relevant knowledge records, and the response generator produces the final system response using the retrieved knowledge records and the dialogue context.
Comprehensive experiments show that Q-TOD consistently outperforms all baselines on three active task-oriented dialogue datasets, achieving a new state-of-the-art performance.

\section*{Limitations}

It is known that large-scale generation models are hindered by inference inefficiency.
In Q-TOD, the query generator and the response generator are invoked sequentially, which inevitably increases the inference latency.
Besides, similar to the previous works~\citep{roller-etal-2021-recipes, shuster-etal-2021-retrieval-augmentation}, Q-TOD also suffers from the knowledge hallucination problem.
These findings suggest that further research should be undertaken to explore more efficient inference strategy and high-fidelity knowledge-grounded response generation.

\section*{Acknowledgements}

We would like to thank the anonymous reviewers for their insightful and constructive comments.
We also thank Wen Huang, Shiwei Huang, and Jingzhou He for their help in resource coordination.
This work was supported by the National Key Research and Development Project of China (No. 2018AAA0101900).

\bibliography{anthology,custom}
\bibliographystyle{acl_natbib}

\clearpage

\appendix

\section{Hyper Parameters}
\label{sec:appendix-hyper-parameters}

The settings of the hyper parameters used in our experiments are summarized in Table~\ref{tab:hyper-parameters}.

\begin{table}[htp]
    \renewcommand\arraystretch{1.2}
    \centering
    \small
    \begin{tabular}{lccc}
        \hline
        \textbf{Parameters} & \textbf{SMD} & \textbf{CamRest} & \textbf{MWOZ} \\
        \hline
        \hline
        Optimizer & AdamW & AdamW & AdamW \\
        LR Scheduler & Noam & Noam & Noam \\
        LR & 3e-5 & 1e-5 & 3e-5 \\
        Batch Size & 128 & 128 & 128 \\
        Epoch & 50 & 50 & 50 \\
        Beam Size & 4 & 4 & 4 \\
        Input Length & 1024 & 1024 & 1024 \\
        Output Length & 128 & 128 & 128 \\
        \hline
    \end{tabular}
    \caption{Hyper parameters used for SMD, CamRest, and MWOZ.}
    \label{tab:hyper-parameters}
\end{table}

\section{Domain-wise Performance}
\label{sec:appendix-domain-wise-performance}

For detailed analysis, we also provide the performance of Q-TOD on each domain of SMD and MWOZ in Table~\ref{tab:domain-wise-results}.

\begin{table}[htp]
    \renewcommand\arraystretch{1.1}
    \centering
    \begin{tabular}{lcc}
        \hline
        \textbf{Domain} & \textbf{T5-Large} & \textbf{T5-3B} \\
        \hline
        \hline
        SMD Schedule  & 81.42 & 84.22 \\
        SMD Navigate & 62.91 & 62.72 \\
        SMD Weather & 69.18 & 73.17 \\
        \hline
        MWOZ Hotel & 45.25 & 46.71 \\
        MWOZ Attraction & 54.81 & 62.68 \\
        MWOZ Restaurant & 55.78 & 58.90 \\
        \hline
    \end{tabular}
    \caption{Domain-wise Entity F1 of Q-TOD on SMD and MWOZ.}
    \label{tab:domain-wise-results}
\end{table}

\section{Exploration on the Number of Retrieved Knowledge Records}
\label{sec:appendix-exploration-on-the-number-of-retrieved-knowledge-records}

In our experiments, the knowledge retriever outputs top-$n$ relevant knowledge records for response generation.
To explore the effect of adjusting the number of retrieved knowledge records, we evaluate Q-TOD in three different top-$n$ settings on the validation set of SMD.
In Table~\ref{tab:exploration-on-the-number-of-retrieved-knowledge-records}, it is observed that the model using top-3 retrieved knowledge records achieves the best Entity F1 on both T5-Large and T5-3B.
This suggests that a small number of $n$ might be inadequate to cover the necessary knowledge records for response generation.
In contrast, a large number of $n$ would inevitably introduce more noisy knowledge records and increase the difficulty of knowledge utilization in response generation.

\begin{table}[htp]
    \renewcommand\arraystretch{1.1}
    \centering
    \begin{tabular}{lccc}
        \hline
        \textbf{Model} & \textbf{Top-1} & \textbf{Top-3} & \textbf{Top-5} \\
        \hline
        \hline
        Q-TOD (T5-Large) & 68.98 & \textbf{70.77} & 70.36 \\
        Q-TOD (T5-3B) & 70.29 & \textbf{72.34} & 71.91 \\
        \hline
    \end{tabular}
    \caption{Performance of Q-TOD with top-$n$ retrieved knowledge records on the validation set of SMD.}
    \label{tab:exploration-on-the-number-of-retrieved-knowledge-records}
\end{table}

\section{Cross-domain Dataset Construction}
\label{sec:appendix-cross-domain-dataset-construction}

We construct two cross-domain dialogue datasets by merging single-domain dialogue sessions from SMD, CamRest, and MWOZ.
Table~\ref{tab:cross-domain-dataset-construction} describes the construction details for these two datasets.
For instance, in \texttt{SMD+CamRest}, two original dialogue sessions are merged into a cross-domain session, where one is from SMD and the other is from CamRest.
Especially, these sessions are concatenated in a random order.
The two cross-domain datasets both contain 600 dialogue sessions, in which the partition of train/validation/test is 400/100/100.

\begin{table}[htp]
    \renewcommand\arraystretch{1.3}
    \centering
    \scriptsize
    \begin{tabular}{lccc}
        \hline
        ~ & \textbf{SMD+CamRest} & \textbf{SMD+CamRest+MWOZ} \\
        \hline
        \hline
        \multirow{3}*{session A} & navigate (SMD) & navigate (SMD) \\
         ~ & schedule (SMD) &  schedule (SMD) \\
         ~ & weather (SMD) & ~ \\
        \hline
        \multirow{2}*{session B} & restaurant (CamRest) & restaurant (CamRest) \\
        ~ & ~ & restaurant (MWOZ) \\
        \hline
        \multirow{3}*{session C} & \multirow{3}*{-} & weather (SMD) \\
        ~ & ~ & attraction (MWOZ) \\
        ~ & ~ & hotel (MWOZ) \\
        \hline
    \end{tabular}
    \caption{Detailed cross-domain dataset construction.}
    \label{tab:cross-domain-dataset-construction}
\end{table}

\end{document}